\documentclass{elsarticle}

\usepackage{url}
\usepackage{booktabs}
\usepackage{amsmath}
\usepackage{multirow}
\usepackage[misc]{ifsym}
\usepackage{amsmath}
\usepackage{graphicx}
\usepackage{csquotes}
\usepackage{amsfonts}
\usepackage{multicol}
\usepackage{tabularx}
\usepackage{appendix}
\begin{document}
\begin{frontmatter}

\title{EdgeCNN: Convolutional Neural Network Classification Model with small inputs for Edge Computing}
\author[1]{Shunzhi Yang}
\author[1]{Zheng Gong\corref{correspondingauthor}}
\cortext[correspondingauthor]{Corresponding author}
\ead{cis.gong@gmail.com}
\author[1]{Kai Ye}
\author[1]{Yungen Wei}
\author[2]{Zheng Huang}
\author[1]{Zhenhua~Huang}
\address[1]{School of Computer Science, South China Normal University, Guangzhou, 510665, China.}
\address[2]{School of Information Security Engineering, Shanghai Jiaotong University, Shanghai, 200240, China.}

\begin{abstract}
With the development of Internet of Things (IoT), data is increasingly appearing on the edge of the network. Processing tasks on the edge of the network can effectively solve the problems of personal privacy leaks and server overload. As a result, it has attracted a great deal of attention and made substantial progress. This progress includes efficient convolutional neural network (CNN) models such as MobileNet and ShuffleNet. However, all of these networks appear as a common network model and they usually need to identify multiple targets when applied. So the size of the input is very large. In some specific cases, only the target needs to be classified. Therefore, a small input network can be designed to reduce computation. In addition, other efficient neural network models are primarily designed for mobile phones. Mobile phones have faster memory access, which allows them to use group convolution. In particular, this paper finds that the recently widely used group convolution is not suitable for devices with very slow memory access. Therefore, the EdgeCNN of this paper is designed for edge computing devices with low memory access speed and low computing resources. EdgeCNN has been run successfully on the Raspberry Pi 3B+ at a speed of 1.37 frames per second. The accuracy of facial expression classification for the FER-2013 and RAF-DB datasets outperforms other proposed networks that are compatible with the Raspberry Pi 3B+. The implementation of EdgeCNN is available at https://github.com/yangshunzhi1994/EdgeCNN
\end{abstract}


\end{frontmatter}

\section{Introduction}
It is estimated that there will be 18 billion IoT devices by 2022 \cite{52}. The trend in IoT is to distribute and move computing from centralized cloud devices to edge devices which are closer to data sources \cite{53}. The computing made on the data source are also called edge computing. We define edge computing as a way to process data on resource-constrained terminal devices. Edge computing has the potential to address the concerns of response time requirements, bandwidth cost saving, connectivity, data safety and privacy, and server overload \cite{47,48,49,55,57}. However, many edge computing devices are characterized by limited computational and energy resources \cite{56,58}. At the same time, many tasks have strict requirements on computational cost. This means that running tasks that require high computing power on edge computing devices is very challenging. Currently, there are four main ways to solve this problem.

The first method is to perform data encryption on the edge computing device before uploading it to a more powerful server \cite{10,16,17,46,54}. This approach introduces a slight computational overhead on the edge computing device and moves most workload to the server, which has sufficient computing resources. Moreover, the use of data encryption protects the privacy of users. However, the data transmission also increases the pressure on network bandwidth and introduces the possibility of network latency. In particular, when the network is inaccessible, the edge computing device cannot operate normally.

The second method is to run the task directly on the edge computing device using traditional machine learning methods, such as those in the field of facial expression  classification \cite{1,2,3,4,5,6}. This is feasible because traditional machine learning methods require relatively little computing power and memory \cite{14}. This means that they are relatively easy to run in resource-constrained edge computing devices. However, the feature attributes extracted by these traditional machine learning methods rely on manual design. Manually designed features are not very robust to the diversity of targets \cite{11}. This problem also occurs in the field of facial expression recognition. As described in \cite{12}, the features extracted by these traditional methods are generally low-level features. They are often effective in specific small datasets, but cannot sufficiently adjust to identifying new facial expression images.

The third method is to use multiple edge computing devices to perform a task together \cite{53,60}. This method needs to focus on computation and communication resource allocation and data sharing issues. The third method is particularly useful for tasks that require data to be collected from multiple edge computing devices for further processing. But for tasks that only require a single data source, this resulting in increased network load and  transmission latency \cite{61}.

The fourth method is to use an edge computing device to run deep learning methods, such as CNNs. The success of CNNs comes from their rich representation capacity and powerful generalization ability for image recognition \cite{13}. The features extracted by CNNs are advanced and are highly robust to multi-style changes of targets. Unlike traditional machine learning methods, deep learning-based methods, using CNNs, extract optimal features with the desired characteristics directly from data \cite{14}. Moreover, they also reduce the dependence on physics-based models and other pre-processing techniques \cite{15}. The deep learning method is usually considered the most promising. However, deep learning requires high computing power, which means that running deep learning on edge computing devices is very challenging. Compute and memory demands of deep learning methods must be addressed to make them useful on IoT endnodes \cite{59}.

To be able to run CNN models on edge computing devices, some efficient CNN models have been proposed \cite{29,30,31,36,37,38,39,27,43,9,8}. However, all of these networks appear as a common network model  and they usually need to identify multiple targets when applied. Therefore, the size of the input to the network model is very large. This leads to a large number of parameters and requires more computational power (FLOPs) in a given situation. Target recognition requires finding and classifying targets at unknown locations. However, the target classification only needs to classify the targets at known locations. In some specific cases, only the target needs to be classified. The input size of those common network models is generally 224 $\times$ 224, which contains multiple targets and related background. If these networks are applied to target classifications, this can result in wasted resources and possible poor performance. Therefore, this paper proposes a new CNN model based on edge computing devices named EdgeCNN. EdgeCNN uses an input size of 44 $\times$ 44 because this size is already adequately categorized. If the input size of the network is reduced, the learning characteristics are rough relative to a network with large inputs. However, classification is based on learning the similarities of individual features. 

What we need to know is that the last layer of the current deep learning classification network is a fully connected layer. Take the example of identifying whether the target in the picture is "human". The "head", "body", etc., represent the feature maps learned by the fully connected layer. EdgeCNN's fully connected layer learns relatively large feature maps such as "head" and "body", while the fully connected layer in the 224 $\times$ 224 network learns the ears, eyes, etc. of the "head". The fully connected layer is scored based on the similarity of these learned features. This means that as long as the network is well designed, EdgeCNN's classification doesnot result in low accuracy due to small inputs. In addition, since the learning features are relatively large, the fully connected layer requires only a few dimensional features. This not only reduces complexity, but also allows the network to focus on learning relatively large features.

To the best of our knowledge, this is the first paper discussing a network with small inputs for target classification. Therefore, this paper can compare the performance of networks with small inputs on the facial expression task. The recognition of facial expressions on edge computing devices has high practical value, such as detecting security risks in real time through cameras. In addition, we found that these efficient network models with small inputs have low accuracy on facial expression datasets. More importantly, the group convolution of other efficient network models is not suitable for edge computing devices with low memory access speeds. Therefore, we propose EdgeCNN in this paper, which runs on edge computing devices and has small inputs. The feasibility of EdgeCNN is demonstrated on the classification of facial expressions. The contributions of this paper are as follows:

\begin{enumerate}
\item We propose EdgeCNN, based on edge computing devices with small inputs. The accuracy of EdgeCNN for facial expression classification exceeds that of other proposed networks with small inputs.
\item We showed in the experiment that group convolution is not optimal in all cases. Group convolution is slower on edge computing devices with slower memory access than without group convolution. The actual running speed of the EdgeCNN model without group convolution on the edge computing device basically meets the actual needs.
\item We run the EdgeCNN model on Raspberry Pi 3B+ to prove that facial expressions can be classified and recognized by using deep learning on edge computing devices. 
\end{enumerate}

The remainder of this paper is as follows: Section 2 briefly introduces the principles of ResNet \cite{22}, DenseNet \cite{23}, and other efficient networks and explains the superiority of the DenseNet principle on resource-constrained devices. Section 3 introduces the design of EdgeCNN in detail. Section 4 shows a performance comparison between EdgeCNN and other efficient networks with small inputs on the facial expression datasets. Section 5 is a summary of this paper.

\section{Related Work}
\label{sec:1}
Since the advent  of the first CNN model in 1989 \cite{18}, the main feature extraction networks that have been developed include LeNet-5 \cite{18}, AlexNet \cite{19}, VGG \cite{20}, GoogLeNet \cite{21}, ResNet \cite{22}, and DenseNet \cite{23}. Among these, the best and most widely used are ResNet and DenseNet because these two networks are very deep compared with previous networks. Researchers have shown that a deeper feature extraction network can learn more advanced features and can classify better. Therefore, we reference  the theoretical methods of these two feature extraction networks to design an efficient network suitable for facial expression classification.

In a general CNN, the output $X_{L}$ at the $L$th layer is:
\begin{equation}
X_{L} = H_{L}(X_{L-1})
\end{equation}

In ResNet, the output $X_{L}$ at the $L$th layer is:
\begin{equation}
X_{L} = H_{L}(X_{L-1}) + X_{L-1}
\end{equation}

In DenseNet, the output $X_{L}$ at the $L$th layer is:
\begin{equation}
X_{L} = H_{L}([X_{0},X_{1 },...,X_{L-1}])
\end{equation}

Among these, $H_{L}(\cdot)$  is a composite function that represents a combined operation. It may include a series of operations such as batch normalization \cite{25}, convolution, and so on.

The feature of the $L$th layer of ResNet comes only from the features of the previous layer and the features obtained by the composite operation of the $L$th layer. The network depth is continuously enhanced by the identity mapping of skip connections to obtain better network performance. From another perspective, ResNet continuously abstracts features through identity mapping to obtain more advanced features. Consequently, there will be many transition features in the middle of the network, which need to be saved in memory when the program is running. Therefore, ResNet  is an unwise basis for the design of an efficient network  on a resource-constrained device.

The feature of the $L$th layer of DenseNet is derived  from all previous layers, which improves the performance of the model through feature reuse. DenseNet can be divided into multiple dense blocks, and the feature maps of all dense blocks have the same size. Each layer's feature map in a dense block is passed to all subsequent layers in the block. The last layer of feature mapping in each dense block is used as input for all subsequent blocks. Thus, each layer in DenseNet generates only a small number of features, but after connection, the last layer of the network can obtain many feature maps. In the extreme case, only one feature is learned in each layer of the network; naturally, this is very difficult to achieve. Consequently, DenseNet greatly reduces the number of intermediate transition features. This is equivalent to reducing the consumption of memory resources, which is very important for resource-constrained devices. Moreover, we can also exploit the shared memory allocation strategy in Ref. \cite{28} to further reduce the use of memory.

However, substantial computing power is required to implement an efficient network with low memory consumption and high accuracy by using DenseNet's principles solely. For example, DenseNet (growthRate=12, depth=40) of 44 $\times$ 44-pixel input image size executes 166.9 MFLOPs. The current main methods of improving network efficiency are group convolution \cite{31,38,39}, depthwise separable convolution \cite{36,30,31,37}, and grouping \cite{31,37,38,39,40} to reduce the computational power of the network model. In fact, depthwise separable convolution is a special case of group convolution. Therefore, the problem of network computing power is currently solved mainly by the two mainstream methods of group convolution and grouping. We need to pay attention to the fact that group convolution can reduce the amount of computation, but it will increase the cost of memory access \cite{37}. Therefore, in this paper, a network using group convolution (EdgeCNN-G) and a network without group convolution (EdgeCNN) are designed. The grouping method is to solve the problem of information flow communication between group and group after grouping. Currently, there are three methods: the channel shuffle of ShuffleNet \cite{31,37}, the learned group convolution of CondenseNet \cite{40}, and the interleaved group convolutions of IGCV \cite{38,39}. In the channel shuffle and interleaved  group convolution methods, assigning these features to disjoint groups can hinder the effective reuse of features in the DenseNet network \cite{40}. Therefore, we use the learned group convolution of CondenseNet, which allows the network to automatically discover good connectivity patterns in learning to reduce the computational power.

CondenseNet is an upgraded version of the DenseNet network. However, the computing power required by the CondenseNet network is still very large. For an input image size of 44 $\times$ 44 pixels, the CondenseNet network performs 74.11 MFLOPs. We directly reduced parameters of CondenseNet, such as the growth rate, but found that the accuracy on datasets for facial expression classification and recognition was not good. There is a no-free-lunch theorem in machine learning, which states that there is no algorithm that can be optimal on all sets of problems \cite{32}. The no-free-lunch theorem tells us that we need to select and construct appropriate deep learning models based on specific sets of problems and specific datasets.

To further reduce the parameters and computational power required by the network model based on edge calculation, we adopt DenseNet's feature reuse method and learned group convolution to design an EdgeCNN-G model with small inputs. In order to compare the effects of group convolution, this paper only uses DenseNet's feature reuse method to design an EdgeCNN model with small inputs. 

First, we use the dlib \cite{50} library to capture faces. The implementation of this part can be viewed in our other project implementation$\footnote{ \url{https://github.com/tobysunx/face_recognition} }$. It captures a face at random times of 0.6 to 3 seconds. Capturing faces from the video is very smooth and not stuck. In this paper, the image of the face captured by the dlib library is compressed to a uniform size of 44 $\times$ 44 pixels. This is because the dlib face detector recognizes faces very accurately \cite{35,7}. In addition, the dlib \cite{50} library consumes much less computing power than  the deep convolutional network model. A face image of 44 $\times$ 44 pixels is also classified sufficiently well. 

Next, the CNN model is used for feature extraction and classification. This is equivalent to translating a multi-target recognition problem to a single-target classification problem. Therefore, it decreases  the size of the network input, reducing the number of parameters and reducing the computational power. 

Although this paper demonstrates the feasibility of EdgeCNN and EdgeCNN-G on facial expression datasets, it can also be applied to the classification and recognition of other targets. However, to use it, it is necessary to find the location of the target first. For example, the cascading \cite{45} method can be used. First, a network input image of 224 $\times$ 224 pixels is used. Second, the RPN \cite{44} method is used to find the target. Finally, EdgeCNN and EdgeCNN-G is employed  to analyze the specific categories.

\begin{table}
\caption{Extract low-level layer feature }
\label{table:1}
\begin{center}
\scalebox{0.91}{
\begin{tabular}{lll}
\hline\noalign{\smallskip}
& DenseNet & EdgeCNN, EdgeCNN-G  \\
\noalign{\smallskip}\hline\noalign{\smallskip}
Image size of input & $112\times112\times3$ & $44\times44\times3$ \\
\noalign{\smallskip}\hline\noalign{\smallskip}
Convolutional block & $7\times7$conv,strides=2 & $3\times3$conv,pad=1,bias=True \\
\noalign{\smallskip}\hline
\end{tabular}}
\end{center}
\end{table}

\section{Architecture Design}
\label{sec:1}
One of the differences between the original DenseNet network and EdgeCNN and EdgeCNN-G is the extraction of the low-level layer features, as shown in Table \ref{table:1}. This part of the network  contains many details. We need to extract as many features as possible. Because the features of the low-level layer learning are not enough, it is difficult to extract effective advanced features in the lower layers. However, a larger network is not necessarily better: this would waste memory, time, and computing power. Considering that the appearance of facial expressions occupies  a certain area, it is necessary to appropriately increase the size of the convolution kernel in the low-level layer features to adapt to the size of the facial expressions. Increasing the size of the convolution kernel enlarges the receptive field. It can learn more detailed information, and there are more parameters followed. In the low-level layer feature of DenseNet, a $7 \times 7 $ convolutional layer is used, which requires very substantial computing power. This paper uses a $3\times 3$ convolutional layer to obtain the low-level layer features of the image, and this is sufficient for networks with small inputs.

\begin{table}
\caption{EdgeCNN's EdgeBlock }
\label{table:2}
\begin{center}
\scalebox{0.81}{
\begin{tabular}{lll}
\hline\noalign{\smallskip}
 & DenseNet & EdgeCNN  \\
\noalign{\smallskip}\hline\noalign{\smallskip}
\multirow{2}{*}{Convolutional}& BN & 3$\times$3 Conv,out\_channels=4$\times$growth\_rate\\
\multirow{2}{*}{block 1} & ReLU & BN\\
& 1$\times$1 conv & ReLU\\
\noalign{\smallskip}\hline\noalign{\smallskip}
\multirow{2}{*}{Convolutional}& BN &\multirow{2}{*}{3$\times$3 Conv,out\_channels=1$\times$growth\_rate}\\
\multirow{2}{*}{block 2} & ReLU & \multirow{2}{*}{BN}\\
& 3$\times$3 conv & \\
\noalign{\smallskip}\hline
\end{tabular}}
\end{center}
\end{table}

\begin{table}
\caption{EdgeCNN-G's EdgeBlock }
\label{table:3}
\begin{center}
\scalebox{0.81}{
\begin{tabular}{lll}
\hline\noalign{\smallskip}
 & DenseNet & EdgeCNN-G  \\
\noalign{\smallskip}\hline\noalign{\smallskip}
\multirow{2}{*}{Convolutional}& BN & 3$\times$3 L-Conv,G=4,C=4,out\_channels=4$\times$growth\_rate\\
\multirow{2}{*}{block 1} & ReLU & BN\\
& 1$\times$1 conv & ReLU\\
\noalign{\smallskip}\hline\noalign{\smallskip}
\multirow{2}{*}{Convolutional}& BN &\multirow{2}{*}{3$\times$3 L-Conv,G=8,C=8,out\_channels=1$\times$growth\_rate}\\
\multirow{2}{*}{block 2} & ReLU & \multirow{2}{*}{BN}\\
& 3$\times$3 conv & \\
\noalign{\smallskip}\hline
\end{tabular}}
\end{center}
\end{table}

\begin{table}
\caption{Architecture of EdgeCNN}
\label{table:4}
\begin{center}
\scalebox{0.90}{
\begin{tabular}{lll}
\hline\noalign{\smallskip}
Layer & Operator & Output  \\
\noalign{\smallskip}\hline\noalign{\smallskip}
Convolution & 3$\times$3conv,pad=1,bias=True & 44$\times$44$\times$32 \\
\noalign{\smallskip}\hline\noalign{\smallskip}
Pooling & 3$\times$3max pool,stride=2 & 22$\times$22$\times$32 \\
\noalign{\smallskip}\hline\noalign{\smallskip}
EdgeCNN's EdgeBlock (1) &  $\begin{bmatrix}  3\times3 $conv$ \\ 3\times3 $conv$ \end{bmatrix} \times 4$&  22$\times$22$\times$64 \\
\noalign{\smallskip}\hline\noalign{\smallskip}
Transition Layer (1) & 2$\times$2average pool,stride=2 & 11$\times$11$\times$64\\
\noalign{\smallskip}\hline\noalign{\smallskip}
EdgeCNN's EdgeBlock (2) &  $\begin{bmatrix}  3\times3 $conv$ \\ 3\times3 $conv$ \end{bmatrix}\times 4$ &  11$\times$11$\times$96 \\
\noalign{\smallskip}\hline\noalign{\smallskip}
Transition Layer (2) & 2$\times$2average pool,stride=2 & 5$\times$5$\times$96\\
\noalign{\smallskip}\hline\noalign{\smallskip}
EdgeCNN's EdgeBlock (3) &  $\begin{bmatrix}  3\times3 $conv$ \\ 3\times3 $conv$ \end{bmatrix}\times 7 $ &  5$\times$5$\times$152 \\
\noalign{\smallskip}\hline\noalign{\smallskip}
Classification & 5$\times$5global average pool & \multirow{2}*{1$\times$1$\times$152 }\\
Layer & 152D fully-connected, softmax & ~ \\
\noalign{\smallskip}\hline
\end{tabular}}
\end{center}
\end{table}

\begin{table}
\caption{Architecture of EdgeCNN-G}
\label{table:5}
\begin{center}
\scalebox{0.90}{
\begin{tabular}{lll}
\hline\noalign{\smallskip}
Layer & Operator & Output  \\
\noalign{\smallskip}\hline\noalign{\smallskip}
Convolution & 3$\times$3conv,pad=1,bias=True & 44$\times$44$\times$32 \\
\noalign{\smallskip}\hline\noalign{\smallskip}
Pooling & 3$\times$3max pool,stride=2 & 22$\times$22$\times$32 \\
\noalign{\smallskip}\hline\noalign{\smallskip}
EdgeCNN-G's EdgeBlock (1) &  $\begin{bmatrix}  3\times3 $L-conv$ \\ 3\times3 $L-conv$ \end{bmatrix} \times 4$&  22$\times$22$\times$64 \\
\noalign{\smallskip}\hline\noalign{\smallskip}
Transition Layer (1) & 2$\times$2average pool,stride=2 & 11$\times$11$\times$64\\
\noalign{\smallskip}\hline\noalign{\smallskip}
EdgeCNN-G's EdgeBlock (2) &  $\begin{bmatrix}  3\times3 $L-conv$ \\ 3\times3 $L-conv$ \end{bmatrix}\times 4$ &  11$\times$11$\times$96 \\
\noalign{\smallskip}\hline\noalign{\smallskip}
Transition Layer (2) & 2$\times$2average pool,stride=2 & 5$\times$5$\times$96\\
\noalign{\smallskip}\hline\noalign{\smallskip}
EdgeCNN-G's EdgeBlock (3) &  $\begin{bmatrix}  3\times3 $L-conv$ \\ 3\times3 $L-conv$ \end{bmatrix}\times 7 $ &  5$\times$5$\times$152 \\
\noalign{\smallskip}\hline\noalign{\smallskip}
Classification & 5$\times$5global average pool & \multirow{2}*{1$\times$1$\times$152 }\\
Layer & 152D fully-connected, softmax & ~ \\
\noalign{\smallskip}\hline
\end{tabular}}
\end{center}
\end{table}

In the composite function $H_{L}(\cdot)$, DenseNet uses the pre-activation mode in Ref. \cite{24}, as shown in convolution block 1 of Table \ref{table:2} and Table \ref{table:3}. That is, the batch normalization layer (BN) \cite{25} is followed by a rectified linear unit (ReLU) \cite{26} and a convolutional layer. However, in an experiment, we found that it is less accurate in networks with small inputs. Like Pelee [27], we use the traditional post-activation mode \cite{25}; that is, the convolutional layer is followed by a batch normalization layer and a ReLU layer. Similarly, we also modified the pre-activation mode in the learned group convolution (L-Conv is used to represent the learned group convolution in Tables \ref{table:3}, where the parameter G represents the number of groups, and C is the condensation factor), replacing it by the post-activation mode, which obtained good performance in the experiment. Inspired by MobileNetV2 \cite{30}, we eliminate  the activation layer in the second convolution block of the bottleneck layer. This prevents nonlinearities from destroying too much information; moreover, it reduces  element-wise operations \cite{37}. This paper refers to the structure modified from the bottleneck layer of DenseNet as EdgeBlock. The difference between the EdgeCNN and the EdgeCNN-G model is only whether the learned group convolution is used in the EdgeBlock, as shown in Table \ref{table:2} and Table \ref{table:3} respectively. The overall structure of the EdgeCNN and EdgeCNN-G models is shown in Table \ref{table:4} and Table \ref{table:5}, respectively.

We use two consecutive $3\times3$ convolutional layer in the EdgeBlock, as shown in Table \ref{table:2} and Table \ref{table:3}. This is because a single $3\times3$ convolutional layer is not sufficient to accommodate the size of the facial expressions. In theory, deep learning networks tend to favor larger convolution kernels without considering computational complexity. This is because a larger receptive field can learn richer features. In addition, we eliminated the $1\times1$ convolutional layer because this layer is less accurate in networks with small inputs and adds extra memory consumption. A detailed introduction is as follows.

First, the $1\times1$ convolutional layer has fewer parameters and lower computational complexity than other efficient networks. The trained model is smaller, which is important in some space-constrained terminal devices. However, in general, the storage space of the edge computing device is still sufficient. For example, the Raspberry Pi 3B+ uses an SD card for storage; SD cards are cheap compared with smart devices and have ample storage space. Therefore, the $1\times1$ convolutional layer can be ignored.

Second, in other efficient networks, a large number of $1\times1$ convolutional layers is used  because they can serve to reduce the number of input data channels. The $1\times1$ convolutional layer greatly reduces the computational power required to improve calculation efficiency while ensuring accuracy. Although the $1\times1$ convolutional layer reduces the number of parameters, adding additional layers will consume more memory. When running a convolutional neural network, in addition to the parameters, the output of each network layer also needs to occupy memory. In EdgeCNN and EdgeCNN-G, a $1\times1$ convolutional layer is added to reduce the dimension of the data, but our experiments show that the added extra layer makes the whole network consume the same amount of memory. Memory is important on resource-constrained edge computing devices. In addition, the original $3\times3$ convolutional layer can learn more abundant features than the $1\times1$ convolutional layer. Most importantly, our experiments found that the increased number of $1\times1$ convolutional layers  reduces accuracy in a network with small inputs.

Finally, the use of a $1\times1$ convolutional layer has another purpose: to increase nonlinearity, which allows the neural network to better fit the data. However, adding an active layer or adding a bias to the convolutional layer can also increase nonlinearity. Therefore, we eliminate the $1\times1$ convolutional layer and add bias to all convolutional layers, including learned group convolution, to increase the nonlinear features of the network. Moreover, in EdgeCNN and EdgeCNN-G, all layers, including DenseNet's transition layers, do not use a $1\times1$ convolutional layer. This is because we found that the presence of a 1 $\times$ 1 convolutional layer in a model with small inputs leads to performance degradation.

In summary, the performance of the 1 $\times$ 1 convolutional layer in networks with small inputs is poor, which may be the reason for the poor performance of other efficient networks. The EdgeCNN model designed in this paper is shown in Table \ref{table:4}, with a selected growth rate of 8. After the model is designed, it is the choice of loss function and the setting of optimizer. This paper uses the Softmax Loss function and SGD optimization technology. We use the SGD optimizer to minimize the Softmax Loss function with a learning rate of 1e-2 and a weight decay of 5e-4. At the same time, after 80 epochs, the learning rate dropped by 0.1 after every 5 epochs.

\begin{table}
\caption{Comparison with other efficient networks with small inputs on the Raspberry Pi 3B+ (parameters, FLOPs, speed, memory consumption, and accuracy).}
\label{table:6}
\scalebox{0.68}{
\begin{tabular}{|c|c|c|c|c|c|c|}
\hline
Model& Parameters & FLOPs & Speed & Memory consumed &  FER-2013 Acc& RAF-DB Acc\\
\hline
SqueezeNet \cite{29} & 0.70 MB & 18.12 M & 3.00 FPS & 18.6\% & 68.26\% & 79.98\% \\
\hline
SqueezeNext \cite{43} & 0.56 MB & 9.84 M & 2.16 FPS & 19.0\% & 67.48\% & 78.87\% \\
\hline
MobileNet \cite{36} & 3.06 MB & 27.69 M & 0.21 FPS & 24.2\% & 67.84\% & 82.59\% \\
\hline
MobileNetV2 \cite{30} & 2.12 MB & 16.57 M & 0.25 FPS & 25.0\% & 69.46\% & 81.16\% \\
\hline
MobileNetV3 \cite{8} & 1.18 MB & 4.76 M & 0.92 FPS & 20.8\% & 66.39\% & 79.23\% \\
\hline
ShuffleNet \cite{31} & 0.88 MB & 7.42 M & 0.94 FPS & 20.5\% & 68.51\% & 78.68\% \\
\hline
ShuffleNetV2 \cite{37} & 1.20 MB & 8.14 M & 0.62 FPS & 20.6\% & 67.84\% & 79.88\% \\
\hline
IGCV3 \cite{39} & 2.11 MB & 17.98 M & 0.24 FPS & 30.6\% & 69.90\% & 81.90\% \\
\hline
EfficientNet \cite{9} & 6.82 MB & 28.94 M & 0.13 FPS & 32.8\% & 69.43\% & 81.22\% \\
\hline
MixNet \cite{51} & 5.55MB & 34.92M & 0.17FPS &  35.2\% & 69.82\% & 79.36\% \\
\hline
\textbf{EdgeCNN-G} & \textbf{0.40 MB} & \textbf{2.7 M} & \textbf{0.87 FPS} & \textbf{19.7\%} & \textbf{71.80\%} & \textbf{84.90\%} \\
\hline
\textbf{EdgeCNN} & \textbf{0.40 MB} & \textbf{52.28 M} & \textbf{1.37 FPS} & \textbf{19.1\%} & \textbf{71.80\%} & \textbf{85.13\%} \\
\hline
\end{tabular}}
\end{table} 

\begin{table}
\caption{Network model that cannot run on Raspberry Pi 3B+ (parameters, FLOPs and accuracy are tested on the GPU).}
\label{table:7}
\begin{center}
\scalebox{0.75}{
\begin{tabular}{|c|c|c|c|c|c|c|}
\hline
Model& Parameters & FLOPs & Speed & Memory consumed &  FER-2013 Acc& RAF-DB Acc\\
\hline
IGCV1 \cite{38} & 0.33 MB & 32 M & Null & Null & 70.99\% & 84.28\% \\
\hline
Pelee \cite{27} & 2.0 MB & 14.6 M & Null & Null & 70.57\% & 80.96\% \\
\hline
\end{tabular}}
\end{center}
\end{table} 

\section{Experiments}
\label{sec:1}
Our experiments were designed to run the model trained on the graphics processing unit (GPU) on the Raspberry Pi 3B+ to perform facial expression recognition. Therefore, we needed to compare the real running speed and resource consumption of each model in the final use environment. This is because the Raspberry Pi 3B+ uses a central processing unit (CPU) instead of a GPU. As described in ShuffleNetV2 \cite{37}, the latest cuDNN library is optimized for a $3\times3$ convolutional layer, so it is unlikely that a $3\times3$ convolutional layer is nine times slower than a $1\times1$ convolutional layer. Therefore, this paper compares the actual running speed and memory consumption of each efficient network on the Raspberry Pi 3B+, as shown in Table \ref{table:6} (in which FPS stands for frames per second).

The total memory size of the Raspberry Pi 3B+ is 875 MB, and the input image size for all efficient network models was 44 $\times$ 44 pixels in our experiments. Considering that other efficient networks have not tested the accuracy on facial expression datasets, we also needed to tune the parameters of the other efficient networks. EdgeCNN and other efficient network models with small inputs were evaluated on the FER-2013 \cite{33} and RAF-DB \cite{34} datasets. The accuracy values shown are the highest accuracy achieved after parameter tuning.

The FER2013 \cite{33} dataset contains 28,709 training images, 3,589 validation images, and 3,589 test images, with seven expression labels (anger, disgust, fear, happiness, sadness, surprise, and neutral). The size of each picture is 48 $\times$ 48 pixels. First, we randomly cropped each image to 10 images of 44 $\times$ 44 pixels. Finally, the average score (over all 10 images)  of each facial expression category was calculated by the efficient network to determine the category of the whole picture.

The basic facial expressions datasets of the Real-world Affective Face Data-base (RAF-DB) \cite{34} has 12,271 training images and 3,068 test images and also contains seven expression labels. The size of each picture is 100 $\times$ 100 pixels. First, we resized each picture to 48 $\times$ 48 pixels. Second, we randomly cropped each image to 10 images of 44 $\times$ 44 pixels. Finally, the average score of each facial expression category was calculated by the efficient network to determine the category of the whole picture.

Because there are multiple versions of other efficient networks (for example, SqueezeNext \cite{43} has 1.0-SqNxt-23, 1.0-SqNxt-44, and other models), we only used the models that were detailed in the respective paper or that have good performance. That is, SqueezeNext \cite{43} uses the 1.0-SqNxt-23 model, SqueezeNet \cite{29} uses the 1.0 version of the model, MobileNet \cite{36} and MobileNetV2 \cite{30} use a 1$\times$ expansion factor, MobileNetV3 \cite{8} uses the MobileNetV3-Small model, ShuffleNet \cite{31} uses the ShuffleNet 1$\times$(g=3) model, ShuffleNetV2 \cite{37} uses the ShuffleNet v2 1$\times$ model, IGCV1 \cite{38} uses the IGC-L24M2 model, IGCV3 \cite{39} uses the IGCV3-D 1.0$\times$ model, MixNet \cite{51} uses the MixNet-L model, and EfficientNet \cite{9} uses the EfficientNet-B0 model.

The performance of EdgeCNN, EdgeCNN-G and other efficient networks is shown in Table \ref{table:6} and Table \ref{table:7}. In Table \ref{table:7}, the speed and memory of the IGCV1 and Pelee models are shown as Null, indicating that these two network models are too large to run on the Raspberry Pi 3B+. They can be run by reducing the size of the network parameters, but the performance is even worse than that shown. As shown in Table \ref{table:6}, although the SqueezeNet and SqueezeNext models run very fast, their accuracy is not high. Therefore, EdgeCNN and EdgeCNN-G is very advantageous in terms of accuracy. In addition, EdgeCNN-G outperforms other efficient network models in terms of computing power. Although the computational power is not necessarily related to speed, the computational power required by the model is particularly important if the edge computing device needs to perform multiple tasks.

We can discuss the drawbacks of group convolution from the FLOPs and speeds in Table \ref{table:6}. The group convolution was first proposed from MobileNet \cite{36} and IGCV1 \cite{38}. The SqueezeNet \cite{29} and SqueezeNext \cite{43} models do not use group convolution. The remaining network models use group convolution. Obviously, on the actual running speed of the Raspberry Pi 3B+, the network model using the group convolution is worse than the network model without the group convolution. The reason for this is that group convolution increases the cost of memory access \cite{37}. At the same time, the memory access speed of the Raspberry Pi 3B+ is very slow.

We need to pay attention to the fact that the current efficient network model is mainly designed for embedded devices such as mobile phones. This paper tested the memory access speed of the Raspberry Pi 3B+ to 22.61 MB/s, while the average mobile phone can reach 222.26 MB/s. This means that those embedded devices have 10 times faster memory access than edge computing devices. Therefore, they can use group convolution to design an efficient network model. However, it can be seen from the experiments in Table \ref{table:6} that group convolution is not suitable for edge computing devices with limited computing resources. The use of group convolution requires a comprehensive consideration of the actual network operating environment.

Ref. \cite{37} showed that the ShuffleNetV2 network model is more accurate than MobileNetV2 on the ImageNet 2012 dataset \cite{41,42}. However, as shown in Table \ref{table:6}, in networks with small inputs (as designed in this paper), ShuffleNetV2 has lower accuracy than MobileNetV2 on the FER-2013 and RAF-DB datasets. Therefore, for practical applications, we need to design the appropriate network model based on specific problems.
\section{Conclusion}
\label{sec:1}
This paper proposes the EdgeCNN-G and EdgeCNN model, which can perform deep learning tasks directly on edge computing devices. EdgeCNN-G uses group convolution and EdgeCNN does not use group convolution. It is worth noting that the current efficient network model is mainly used for embedded devices, such as mobile phones. However, the computing resources of most edge computing devices are scarce, such as the Raspberry Pi. The memory access speed of the Raspberry Pi is only 1/10 of that of a average mobile phone. This paper found in the experiment that the group convolution is not applicable to edge computing devices with low memory access speed such as Raspberry Pi. Therefore, the use of group convolution requires a comprehensive consideration of the actual network operating environment.

 EdgeCNN and EdgeCNN-G has been run successfully on the Raspberry Pi 3B+. The EdgeCNN can run on the Raspberry Pi 3B + at 1.37 FPS, which basically meets the actual needs. The parameters and accuracy of the EdgeCNN and EdgeCNN-G model improve on other proposed network models that are compatible with the Raspberry Pi 3B+. However, the EdgeCNN and EdgeCNN-G model has scope for improvement in terms of speed. We tried to use Intel Movidius Neural Compute Stick 2 (NCS 2) $\footnote{OpenVINO\textsuperscript{TM} toolkit: \url{https://docs.openvinotoolkit.org/latest/index.html} }$ on the Raspberry Pi 3B + to increase the speed of EdgeCNN and EdgeCNN-G, but NCS 2 does not support feature detection. However, feature detection is a necessary condition for EdgeCNN and EdgeCNN-G. In addition, the range of edge computing devices is very broad, including, for example, the Advanced RISC Machine (ARM) chip. Currently, the only edge computing devices supported by NCS 2 are Raspberry Pi boards. This means that NCS 2 is not yet widely applicable to edge computing devices. Moreover, we need a ready-made CNN model before using NCS 2. Therefore, we still need to focus on using deep learning methods to improve the speed of task processing on edge computing devices.

\end{document}